\documentclass[sigconf, authorversion, screen]{acmart}

\usepackage{algorithm}
\usepackage{algorithmic}

\AtBeginDocument{%
  \providecommand\BibTeX{{%
    \normalfont B\kern-0.5em{\scshape i\kern-0.25em b}\kern-0.8em\TeX}}}

\def\etal{{\em et al.}}

\newcommand{\BO}[1]{{\boldsymbol{#1}}}

\newcommand{\OP}[1]{{\operatorname{#1}}}

\begin{document}
\renewcommand\footnotetextcopyrightpermission[1]{}
\fancyhead{}

\title[PC$^2$-PU]{
    PC$^2$-PU: Patch Correlation and Point Correlation for Effective Point Cloud Upsampling
}

\author{Chen Long}
\authornotemark[1]
\affiliation{%
    \institution{Wuhan University}
    \city{}
    \country{}
}
\email{chenlong107@whu.edu.cn}

\author{WenXiao Zhang}
\affiliation{%
    \institution{Singapore University of Technology and Design}
    \city{}
    \country{}
}
\email{wenxxiao.zhang@gmail.com}
\authornote{These authors contributed equally.}

\author{Ruihui Li}
\affiliation{%
    \institution{Hunan University}
    \city{}
    \country{}
}
\email{liruihui@hnu.edu.cn}
\authornote{Corresponding author.}

\author{Hao Wang}
\affiliation{%
    \institution{Riemann Lab, Huawei Technologies}
    \city{}
    \country{}
}
\email{wanghao4110@gmail.com}

\author{Zhen Dong}
\affiliation{%
    \institution{Wuhan University}
    \city{}
    \country{}
}
\email{dongzhenwhu@whu.edu.cn}

\author{Bisheng Yang}
\affiliation{%
    \institution{Wuhan University}
    \city{}
    \country{}
}
\email{bshyang@whu.edu.cn}
\renewcommand{\shortauthors}{Chen Long, Wenxiao Zhang.}

\begin{abstract}
    Point cloud upsampling is to densify a sparse point set acquired from 3D sensors, providing a denser representation for the underlying surface.
    Existing methods divide the input points into small patches and upsample each patch separately, however, ignoring the global spatial consistency between patches.
    In this paper, we present a novel method PC$^2$-PU, which explores patch-to-patch and point-to-point correlations for more effective and robust point cloud upsampling.
    Specifically, our network has two appealing designs:
    (i) We take adjacent patches as supplementary inputs to compensate the loss structure information within a single patch and introduce a \emph{Patch Correlation Module} to capture the difference and similarity between patches.
    (ii) After augmenting each patch's geometry, we further introduce a \emph{Point Correlation Module} to reveal the relationship of points inside each patch to maintain the local spatial consistency.
    Extensive experiments on both synthetic and real scanned datasets demonstrate that our method surpasses previous upsampling methods, particularly with the noisy inputs. The code and data are at \url{https://github.com/chenlongwhu/PC2-PU.git}.
\end{abstract}


\maketitle

\section{Introduction}
Point clouds, as a compact representation of 3D surface, provide an effective way for geometry processing. They are widely applied in many fields, such as self-driving cars \cite{selfdrive, zhangr1, zhangr2}, smart cities \cite{batty2012smart}, and robotics \cite{robotics}.
However, due to the inherent limitation of scanning sensors or lighting reflection, raw point clouds acquired from 3D scanners are often sparse, noisy, and occluded.
Hence, such raw data is required to be augmented, before it can be applied
into the downstream works, such as point cloud segmentation \cite{segmentation, Yisegmentation}, surface reconstruction \cite{surface, surface_mm, ZhongHF19, zhangc1, zhangc2, zhangc3}.
To achieve so,
point cloud upsampling is a desirable way to densify the raw point clouds for providing a more faithful description of the underlying surface.

Given a sparse point set, the objective of point cloud upsampling is not limited to producing more points around the inputs.
Also, the input points may be non-uniform and noisy, thus they may not well represent the fine structure of the target objects.
Essentially, the generated points should also be (i) positioned faithfully on the underlying surface and (ii) cover the surface in a uniform manner.
As an inference-based task, these objectives are very challenging to achieve, with the limited information in the sparse input and intrinsic irregularity of 3D point clouds.
\begin{figure}[t]
    \centering
    \includegraphics[width=\columnwidth]{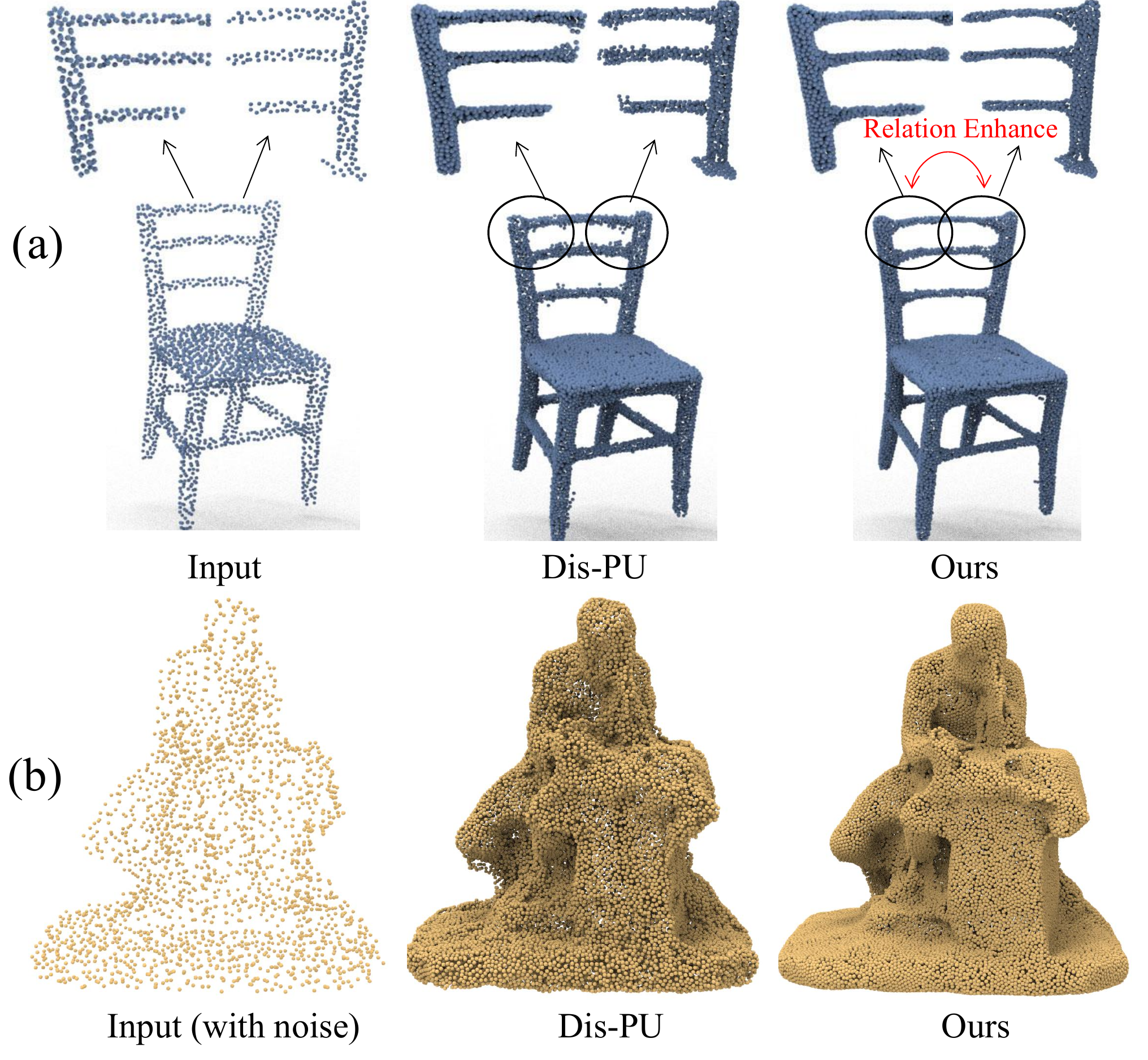} 
    \caption{(a) Given the patches cropped from the input object, existing methods upsamples each patch separately,
        resulting in additional pseudo-points at the boundary of the chair.
        (b) Upsampling results on noisy inputs. Compared with state-of-the-art~\cite{dis-pu}, our method performs more effectively and robustly, by considering both patch-to-patch and point-to-point relationships.}
    \vspace{-0.6cm}
    \label{fig1}
\end{figure}

Generally, previous point cloud upsampling methods can be summarized as optimization-based~\cite{Alexa2003,Lipman2007,Huang2013} and deep-learning-based~\cite{PUNet, 3pu,PUGAN,PUGeo,SSPU-Net,dis-pu}.
In this paper, we follow the line of deep-learning-based methods, which have demonstrated promising upsampling results, with the advent of various neural network architectures for point cloud analysis~\cite{Pointnet, PointNet++}. The general steps taken in existing learning-based upsampling methods are that they first divide the input sparse points into several independent patches, then design a upsampling network to expand the number of points inside each patch, and combine all patches to obtain the final results.

However, current upsampling paradigm has two limitations that restrict its performance.
(i) These methods only \emph{consider the spatial information of each patch separately},
yet ignoring the overall spatial consistency of the target surface.
As shown in Figure \ref{fig1}(a), a single patch may present limited or ambiguous structure information on the boundary. Hence, current methods tend to produce additional pseudo-points in the intersection region between multiple patches. (ii) They focus mainly on a \emph {sparse-to-dense solution with a clean input}, but do not sufficiently consider the local spatial consistency of the generated dense points, the upsampling performance is extremely sensitive to receiving input quality, as shown in Figure \ref{fig1}(b).

To address these problems, we propose to exploit both patch-to-patch and point-to-point relationships during the upsampling process, aiming to maintain the spatial consistency of the target for more efficient and robust upsampling.
The key motivation is that one single patch only presents partial structure information especially on the boundary, thus we propose to incorporate its neighbouring patch during upsampling to compensate the lost information for ensuring a global spatial consistency.
After augmenting each patch's geometry, we maintain the local spatial consistency and restore fine-grained structure by revealing the relationship of points inside each patch.
%

To achieve so, we formulate a novel upsampling framework, named PC$^2$-PU, that consists of a \emph{Patch Correlation Module} (PaCM) and a \emph{Point Correlation Module} (PoCM), accounting for the relations among patches and points, respectively.
Specifically, PaCM captures patch-to-patch information by encoding the similarities and differences in the location of points between adjacent patches, thus expanding the perceptual range of points.
%
%
%
%
PoCM employs a local spatial encoder to encode the relative position information of points in their local field, enabling the network to adjust the generated points to be closer to the underlying surface.
These two modules are jointly optimized and complement with each other to suppress the generation of pseudo-points and dismiss the noise points, achieving a more efficient and robust upsampling performance, as shown in Figure~\ref{fig1} and more results in Section~\ref{Experiments}.

To summarize, the main contributions of our paper include:
\begin{itemize}
    \item We propose to explore both patch-to-patch and point-to-point relationships for more efficient and robust upsampling process.
    \item The designed PaCM captures the similarities and differences between adjacent patches to compensate the lost information of single patch for ensuring global spatial consistency.
    \item We design PoCM to maintain local spatial consistency and restore fine-grained structure by encoding relative position relationships between points.
    \item Extensive experiments show that our method outperforms state-of-the-art upsampling methods on both synthetic and real-scanned datasets, particularly with the noisy inputs.
\end{itemize}

\section{Related Work}
In general, current methods for point cloud upsampling have been mainly classified as optimization-based and learning-based. Optimiz-ation-based approaches often rely heavily on different shape priors \cite{Lipman2007,Alexa2003, Huang2013}, limiting the generalization on diverse 3D structures, particularly when the prior requirements are not satisfied.

Benefiting from the advent of neural networks for 2D image-related tasks, many deep-learning-based methods have been proposed for 3D point cloud analysis~\cite{Pointnet,PointNet++}.
For point cloud upsampling, PU-Net \cite{PUNet} is the pioneering work that is based on PointNet++ \cite{PointNet++}. It divides the input sparse points as small patches and conducts the upsampling operation on each patch.
Yu~\etal~\cite{EC-Net} proposed an edge-aware network EC-Net, which focused on consolidating the edge points.
Later, Wang~\etal~\cite{3pu}proposed a progressive upsampling method MPU, which is motivated by image super-resolution technics.
Li~\etal~\cite{PUGAN} proposed a network structure called PU-GAN and proved that the geometric structure of sparse points is easily lost in the above algorithms.
It introduced the popular GAN \cite{GAN} network into the upsampling task and proposed an innovative up-down-up feature expansion module.
Later, Wu~\etal~\cite{AGGAN} introduced graph convolutional networks to the upsampling task of point clouds and proposed AR-GCN  using adversarial graph loss instead of manually designed loss functions.
Qian~\etal~\cite{pugcn} proposed PU-GCN, arguing that the final quality of point cloud upsampling is heavily dependent on the upsampling modules and feature extractors used therein.
Later, PU-GEO~\cite{PUGeo} was proposed to first generate samples in the 2D domain and then transform them into a 3D domain by a rigorous mathematical formulation.
Li~\etal~\cite{dis-pu} proposed Dis-PU, which consists of two sub-networks that perform the tasks of generation and refinement separately instead of using a single network for upsampling.
Recently, Feng~\etal~\cite{NeuralPoints} proposed a new point cloud representation, which used neural points to get better results.
Due to the difficulty of obtaining dense 3D data, Li et al. proposed a self-supervised point cloud upsampling network SPU-Net \cite{SPU-Net} to generate dense point clouds without using ground truth.
It captures the inherent upsampling pattern of points on the surface of the object to achieve upsampling.
Zhao~\etal~\cite{SSPU-Net} also proposed an unsupervised upsampling network SSPU-Net.
It designed a neighbor expansion unit to upsample the point cloud, and developed a differentiable point cloud rendering unit to render the point cloud into multi-view images to provide additional supervisory signals.
There are also some other novel research directions, such as arbitrary multiplicative point cloud upsampling \cite{meta-pu, MAPU-Net}, multitasking point cloud upsampling \cite{PUI, PUN}, and zero-shot point cloud upsampling \cite{zero-shot}.

\begin{figure*}[t]
    \centering
    \includegraphics[width=\textwidth]{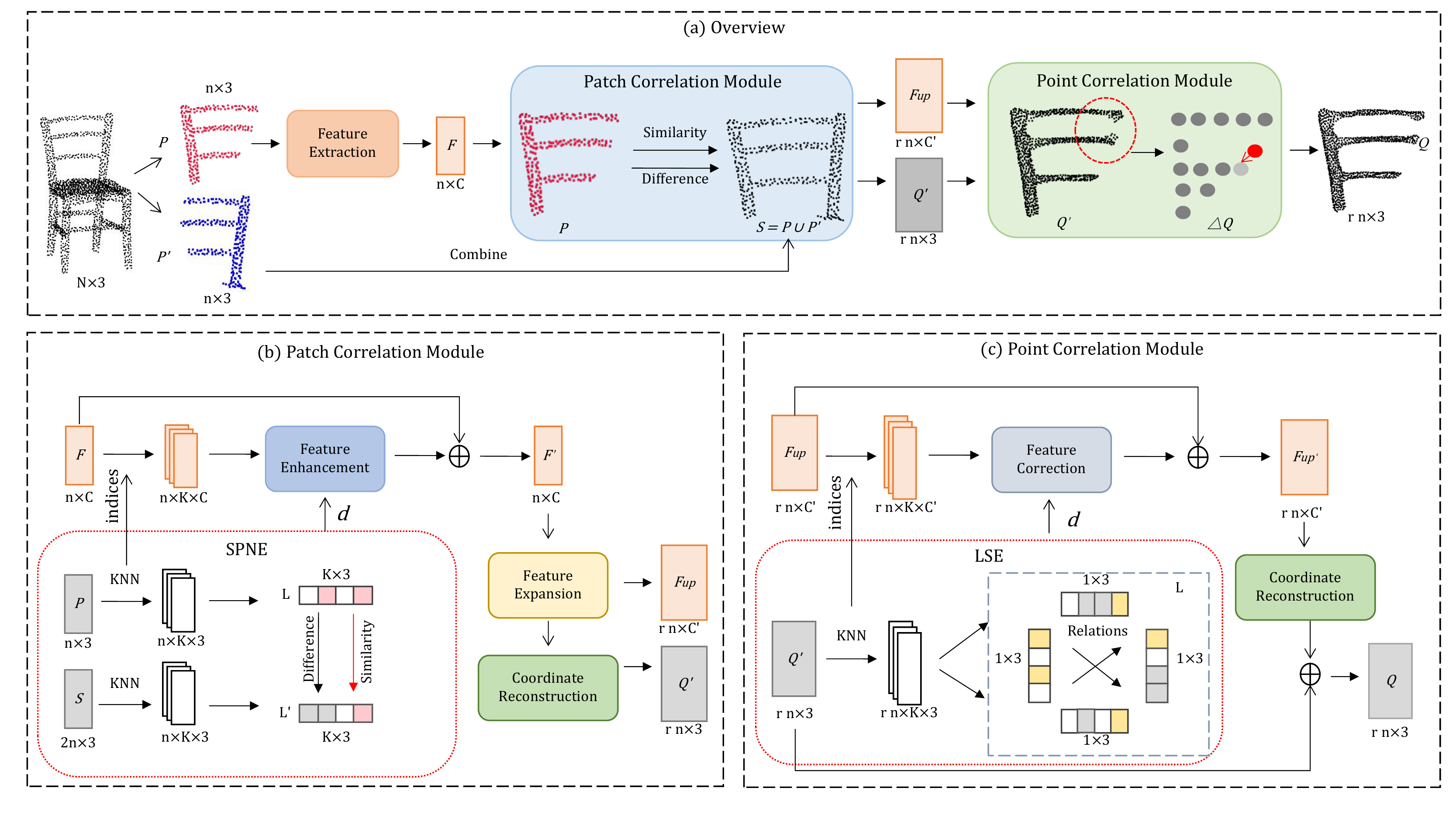} 
    \caption{An illustration of our framework. Given a low-resolution input patch $P$, we firstly obtain its feature maps $\BO F$. Then, we feed the feature $\BO F$, raw patch $P$ and its adjacent patch $P^{\prime}$ into the Patch Correlation Module to combine the information between patches to get a rough point cloud $Q\prime$ and expanded feature map $\BO F_{up}$. Finally, we use the Point Correlation Module to correct the position of the rough point cloud $Q\prime$ and obtain the final high-resolution dense point cloud $Q$.}
    \label{network}
\end{figure*}

Existing methods only consider the spatial structure of each individual patch during upsampling, ignoring the global spatial consistency of the surface, limiting the upsampling ability.
In addition,
existing methods mainly focus on a spare-to-dense solution, but take less account of keeping the local spatial consistency, thus fine-grained structures are easily lost.

\section{Method}
\subsection{Overview}
Given a sparse point set $\mathcal{P}$$=$${\left\{ \BO p_i \right\}}_{i=1}^{N}$ with $N$ points, the goal of upsampling is to generate a dense point cloud $\mathcal{Q}$$=$${\left\{ \BO q_i \right\}}_{i=1}^{rN}$ (where $r$ is the upsampling rate), which provides a more faithful description of the underlying surface within the sparse input points.
    Figure \ref{network}(a) shows the overview of our method, where we also use the patch-wise inputs for network training and each patch $P \in$$\mathbb{R}^{n\times3}$ with $n$ points ($n$$\ll$$N$).
        Different from existing methods that upsample each patch separately, our method picks an adjacent patch $P^{\prime}$ for each input patch $P$ as a supplementary input.
        For training, we randomly select pairs of two adjacent patches as inputs. During testing, we designed a patch selection strategy to select the neighbouring patches with the richest geometric structure as complements, which is detailed in Appendix \ref{a1}.

        Given the input patch $P$, we use the dense feature extraction unit\cite{3pu} as our feature extractor to extract local features $\BO F\in$$\mathbb{R}^{n\times C}$.
    Then we feed the extracted point-wise features $\BO F$, the sparse point cloud $P$ and $P^{\prime}$ through the \emph{Patch Correlation Module}. By capturing the spatial relationship between adjacent patches, we expand the perceptual range of the point and obtain the expanded feature map $\BO F_{up}\in$$\mathbb{R}^{rn\times C^\prime}$ and a rough upsampled point cloud $Q^{\prime}\in$$\mathbb{R}^{rn\times 3}$.
    Then, we feed $Q^{\prime}$ and $\BO F_{up}$ into the \emph{Point Correlation Module}.
    We correct the $Q^{\prime}$ by constructing a local neighbourhood and encoding the relative position information between points, and get the output dense point cloud $Q\in$$\mathbb{R}^{rn\times 3}$.
        Finally, we combined all upsampled patches to obtain the final output $\mathcal{Q}$, as previous methods.
        \subsection{Patch Correlation Module}\label{PRU}

\begin{figure}[h]
            \centering
            \includegraphics[width=\columnwidth]{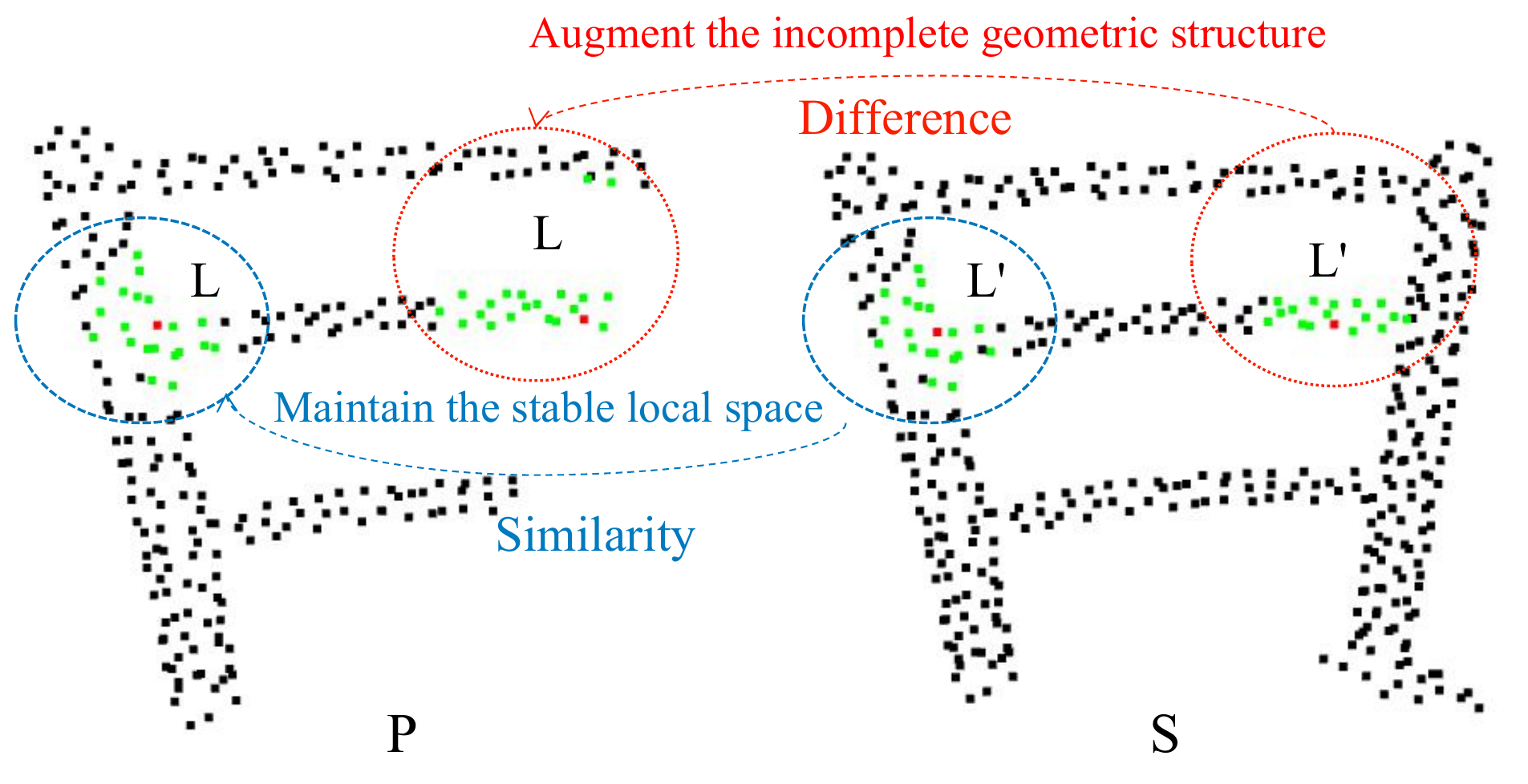} 
            \caption{The red points indicate query points, and the green points are the local neighbors via KNN grouping.}
            \label{pacm_visual}
        \end{figure}

        Figure~\ref{network}(b) depicts the detailed structure of the \emph{Patch Correlation Module (PaCM)}.
        Given the sparse point cloud $P$ and the point-wise feature $\BO F$,
        we first employ KNN grouping on $P$ to search K-nearest neighbors, and group the associated neighbor points together to construct a local neighborhood $\BO L = \{\BO L_1, ..., \BO L_K | \BO L_k \in \mathbb{R}^3 \}$ for each point. At the same time, we use the same nearest neighbor indices to obtain a local neighborhood feature map $\BO X = \{\BO X_1, ..., \BO X_K | \BO X_k \in \mathbb{R}^C \}$ for each point.
        After that, we combine these two adjacent patches $P$ and $P^{\prime}$ to obtain $S = P \cup P^\prime, S \in \mathbb{R}^{2n \times 3}$, and then employ KNN grouping on $S$ to construct a new local neighborhood $\BO L^\prime = \{\BO L^\prime_1, ..., \BO L^\prime_K | \BO L^\prime_k \in \mathbb{R}^3 \}$ for each point inside $P$, as shown in Figure \ref{pacm_visual}. Compared with $\BO L$, $\BO L^{\prime}$ can provide a more complete local reception field, particularly for those points on the boundary of $P$.

        Then, to fuse $\BO L$ and $\BO L^\prime$, we formulate the Feature Enhancement Module as
        \begin{equation}\label{pa_flow}
            \BO F^\prime = \OP {Enhance}(\BO F, \OP {Relation}(\BO L, \BO L^\prime)),
        \end{equation}
        where $\OP {Enhance}(.)$ is a transformer-based architecture \cite{zhao-point}, and $\OP {Relation}(.)$ is an encoder operation to encode the spatial relations between patches, which is used to increase the perceptual range of the points and augment the incomplete geometric structures.

        Specially, we design a SPatial Neighbourhood Encoder (SPNE) as the $\OP {Relation}(.)$ function to explicitly encode the geometric structure information of a query point within the neighborhoods $\BO L$ and $\BO L^\prime$ as follows:
        \begin{equation}\label{SPNE}
            \begin{split}
                \BO d_i^k = (\BO p_i \oplus \BO L_k \oplus (\BO p_i - \BO L_k) \oplus \| \BO p_i - \BO L_k \| \oplus \\
                (\BO L^{\prime}_k - \BO L_k + \BO p_i) \oplus \BO L^{\prime}_k \oplus (\BO p_i - \BO L^{\prime}_k) \oplus \| \BO p_i - \BO L^{\prime}_k\|),
            \end{split}
        \end{equation}
        where $\BO p_i$ is the $i$-th point in $P$. $\BO L_k$ and $\BO L^\prime_k$ are the $k$-th nearest point of $\BO p_i$ within $P$ and $S$, respectively. $\BO d_i$ denotes the position code of $\BO p_i$, and $\BO d_i^k$ is the $k$-th feature channel of $\BO d_i$. $\oplus$ is for the concatenation operator. $\|.\|$ calculates the Euclidean distance between the neighbouring and central points.

        %
        We then leverage the feature $\BO F$ and position code $\BO d$ as the input of $\OP {Enhance}(.)$ function to promote the point-wise feature $\BO F^{\prime}\in$$\mathbb{R}^{n\times C}$ by Eq \ref{point1}, \ref{point2}.
    \begin{equation}\label{point1}
        \BO F_i^\prime = \BO F_i + \sum_{\BO X_k \in \BO X} \gamma(\varphi(\BO F_i) - \psi(\BO X_k) + \delta) \odot (\alpha(\BO X_k) + \delta)
    \end{equation}
    \begin{equation}\label{point2}
        \delta = \OP {Tanh} \circ \OP {MLP}(\BO d_i^k),
    \end{equation}
    where $\BO F_i$ is the feature vector of $\BO p_i$. $\BO X$ is the local neighborhood feature map, and $\BO X_k$ is the feature of the $k$-nearest point to $\BO p_i$. The symbols $\gamma, \varphi, \psi, \alpha$ denote a simple $\OP {MLP}$ layer and $\odot$ denotes element-wise product operation. Here, we choose $\OP {Tanh}$ as the activation function in Eq \ref{point2} in order to preserve and amplify the differences of point positions in adjacent patches.

    To obtain the upsampled feature $\BO F_{up}$,
    we expand the feature channel of $\BO F^{\prime}$ by $r$ times via a graph convolution layer \cite{pugcn}, and then rearrange the feature map to obtain the expanded features $\BO F_{up}\in$$\mathbb{R}^{rn\times C'}$.
    %
    %
    Finally, we use a multi-layer perception ($\OP {MLP}$) as the Coordinate Reconstruction Unit to generate coarse 3D coordinates of dense point cloud $Q^{\prime}$.

    There is an intuitive explanation for the design of PaCM. In our observation, the single-patch methods only consider local incomplete structures, leading the network to generate some pseudo-points on the boundary. To solve this problem, \emph{we augment the incomplete geometric structure by using relations between adjacent patches}. 
    Thus, we first construct the local neighborhood $\BO L$ and $\BO L^{\prime}$ of points within different patches, and then capture the difference and similarity between patches by SPNE, which is sensitive to local spatial structure. This discrepancy between patches represents the missing geometric structure in a single patch, while the similarity represents the stable local space within the patch, as shown in Figure \ref{pacm_visual}. By exploiting the relationship between patches, our method effectively extends the perceived range of points and maintains spatial consistency.

    \textbf{Discussion. Can we get the same results by increasing the patch size?}
    On the one hand, it is hard to choose a proper patch radius to fit the all cases through the entire shape. Taking the case in Figure \ref{fig1}(a), selecting a larger patch radius may also have some incomplete geometric structures. On the other hand, choosing a patch size that is large enough may solve the spatial consistency issue, but it will lose the shape details, increase the number of parameters in the network, and reduce the generalization ability. Therefore, we conduct an experiment in Table \ref{big_patch}, \ref{g_ability} to compare the effectiveness of using adjacent patches and large patch size.

    \subsection{Point Correlation Module}
    Existing networks such as PU-Net \cite{PUNet}, PU-GAN \cite{PUGAN}, Dis-PU \cite{dis-pu} etc., focus mainly on getting a dense point cloud from a clean input, but do not consider the local spatial consistency of the generated points, limiting the robustness of the network. To address this issue, we propose another module named \emph{Point Correlation Module}, as shown in Figure \ref{network} (c).

    \begin{figure}[t]
        \centering
        \includegraphics[width=\columnwidth]{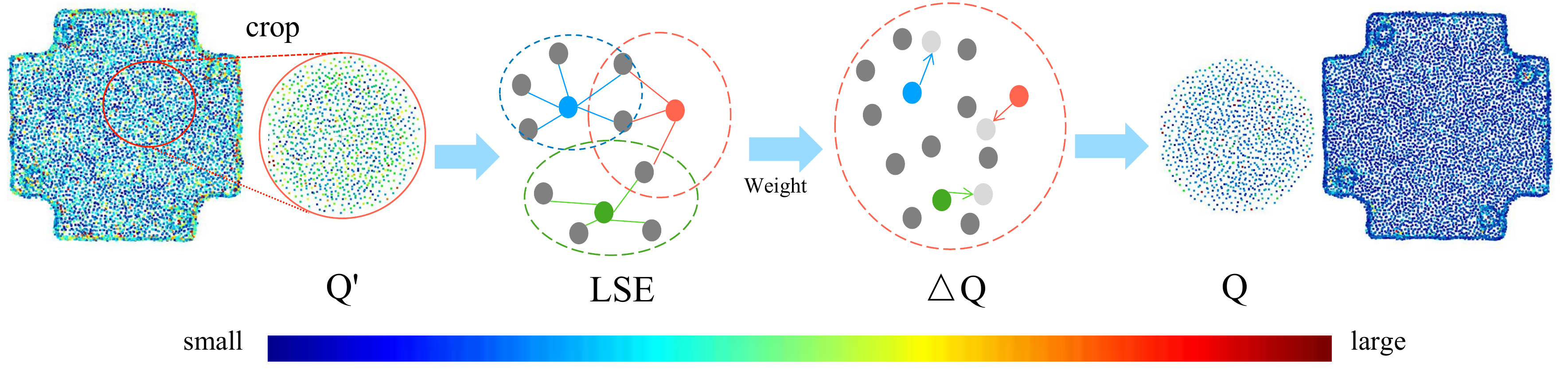} 
        \caption{The dotted circle indicate the neighbourhoods of the query points, and the colors reveal the error distance for each point, and more blue means better results.}
        \label{pocm_visual}
    \end{figure}

    \begin{figure*}[t]
        \centering
        \includegraphics[width=\textwidth]{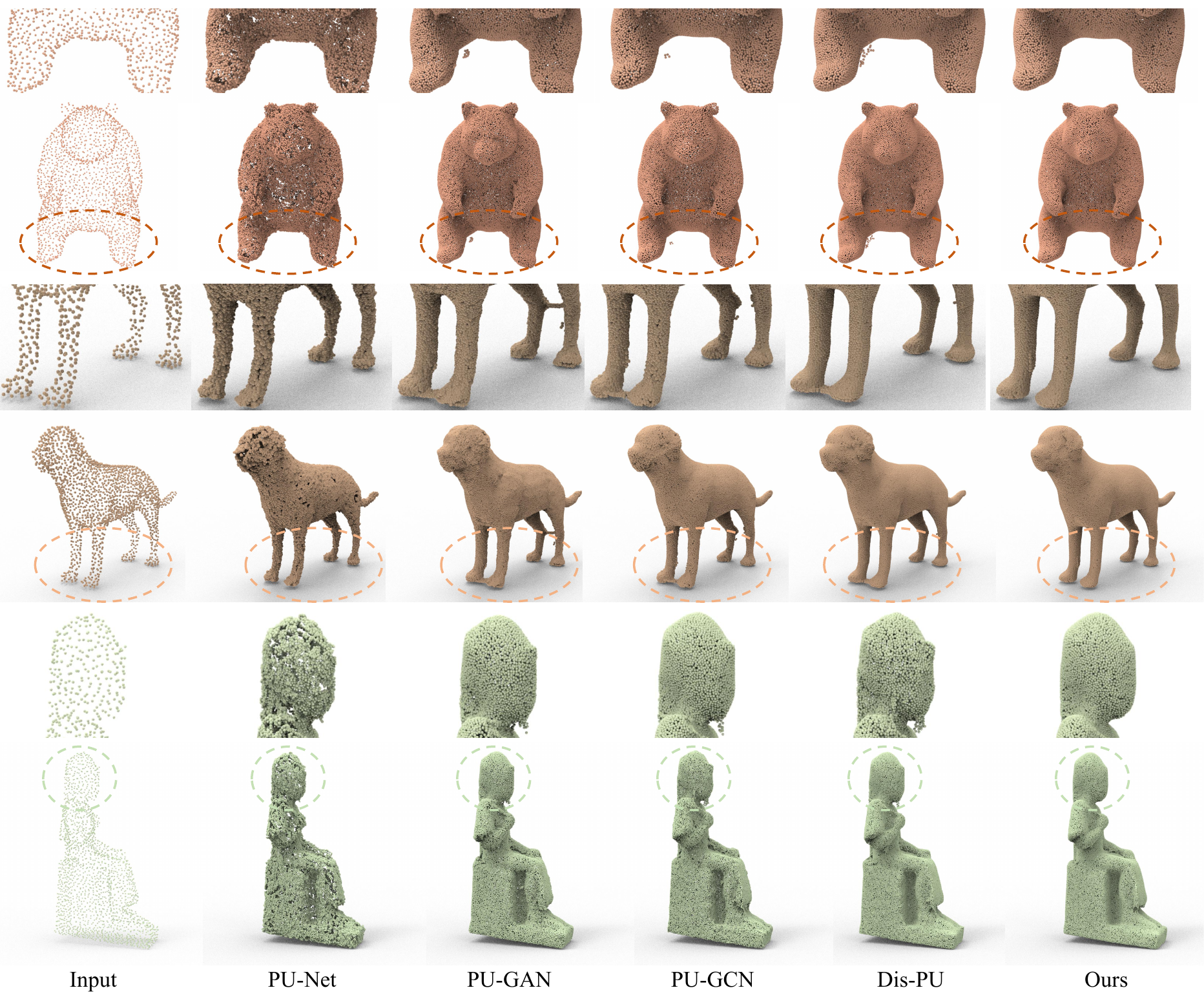} 
        \caption{Comparing upsampling results ($\times16$) from synthetic sparse input (2048 points) using different methods.}
        \label{benchmark}
    \end{figure*}
    \begin{figure*}[t]
        \centering
        \includegraphics[width=\textwidth]{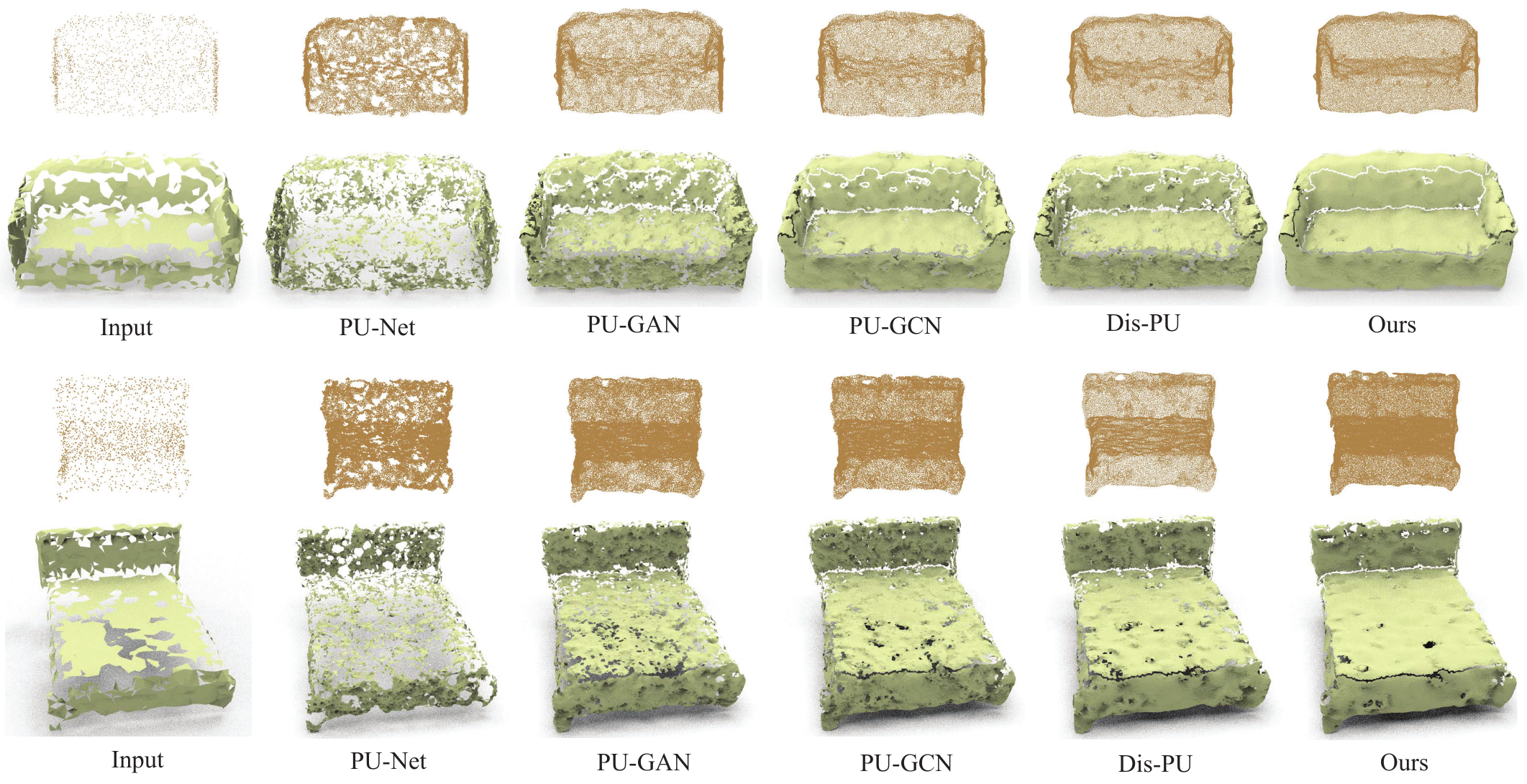} 
        \caption{Comparing the upsampling results ($\times16$) on real-scanned input and the reconstructed 3D meshes. Clearly, our method outperform others in the presence of noisy scanned points, providing a more complete and smooth surface. }
        \label{realscan}
    \end{figure*}

    Given the expanded features $\BO F_{up}$ and the rough point cloud $Q^{\prime}$ as input, we first employ the KNN grouping to generate each point's local neighborhood $\BO L$ and match neighborhood feature map $\BO X$. \emph{It is worth noting that we only consider a single patch in this Module and do not aggregate adjacent patches}. This is because the information between patches is already contained in the inputs $\BO F_{up}$ and $Q^{\prime}$. Furthermore, in this way, the network can focus more on the basic local structure of the 3D shape rather than the entire shape itself, thereby reducing the latent space for estimating the entire 3D geometric representation. Then, we formulate the Feature Correction Module in Eq. \ref{po_flow} to exploit the relationship between the points.
    \begin{equation}\label{po_flow}
        \BO F_{up}^\prime = \OP {Correction}(\BO F_{up}, \OP {Weight}(\OP {Relation}(\BO L))),
    \end{equation}
    where the $\OP {Relation}(.)$ is an encoder operation, $\OP {Weight}(.)$ is a function that calculates the weight of each point, and $\OP {Correction}(.)$ is a transformer-based architecture \cite{zhao-point}.


    Specially, we used a Local Spatial Encoder(LSE) \cite{randla} as the $\OP {Relation}(.)$ function, calculated by Eq \ref{LSE}.
    \begin{equation}\label{LSE}
        \BO d_i^k = (\BO p_i \oplus \BO L_k \oplus (\BO p_i - \BO L_k) \oplus \| \BO p_i - \BO L_k \|).
    \end{equation}
    The definitions of the symbols are the same as those of Eq \ref{SPNE}.
    We feed the position code $\BO d$ and the expanded feature $\BO F_{up}$ through the $\OP {Correction}(.)$ function, which is detailed in Eq. \ref{point3}, \ref{point4}.
    \begin{equation}\label{point3}
        \BO F_{{up}_{i}}^\prime  = \BO F_{{up}_{i}} + \sum_{\BO X_k \in \BO X}\underbrace{\gamma(\varphi(\BO F_{{up}_{i}}) - \psi(\BO X_k) + \delta)}_{\text{Weight(.)}} \odot (\alpha(\BO X_k) + \delta)
    \end{equation}
    \begin{equation}\label{point4}
        \delta = \OP {Relu} \circ \OP {MLP}(\BO d_i^k).
    \end{equation}
    The definitions of the symbols are the same as those of Eq. \ref{point1}, \ref{point2}.
    Different from the \emph{Patch Correlation Module}, we use $\OP {Relu}$ \cite{Relu} as the activation function of the position encoder, which can help the network converge faster. Then, we use the corrected feature $\BO F_{{up}}^\prime$ to regress the offset $\bigtriangleup Q$ for each point through a multi-layer perception structure. Finally we add the predicted offset $\bigtriangleup Q$ to the rough point cloud $Q^{\prime}$ to get the final refined point cloud $Q$.

    The PoCM effectively preserve the local spatial information of the points by encoding the point-to-point correlations, which allows the network to explicitly observe local geometric patterns and thus ultimately facilitate effective learning of complex local structures across the network.
    Moreover, we also use the correlations between points as cues to adjust the weight of each neighboring point, thus correcting the position of the generated points to maintain the local spatial consistency, as shown in Figure \ref{pocm_visual}. Besides, our proposed SPNE and LSE are distance-sensitive, helping the network to distinguish the differences between noise and other points, thus improving the robustness of the network.

    \subsection{Loss function}
    In order to make the final generated points more distributed over the object surface and produce better visual quality, we used the Earth Mover's Distance(EMD)  \cite{fan2017point} as the reconstruction loss function.

    \begin{equation}\label{rec}
        L_{rec} = L_{\OP{EMD}}(Q\prime, \hat{Q}) + \lambda L_{\OP{EMD}}(Q, \hat{Q})
    \end{equation}
    \begin{equation}\label{emd}
        L_{\OP{EMD}}(Q_1, Q_2) = \mathop{min}\limits_{\phi:Q_1 \rightarrow Q_2}\sum_{\BO q_i \in Q}\|\BO q_i - \phi(\BO q_i) \|_2,
    \end{equation}

    where: $\phi:Q_1 \rightarrow Q_2$ is the bijection mapping. $Q\prime$ is the coarse point cloud obtained after the coordinate regression, and $\hat{Q}$ is the ground truth corresponding to the low-resolution input. The $\lambda$ changes as the training rounds increases, to wait for the coarse point cloud generation to be stabilized.

\section{Experiments}\label{Experiments}
\subsection{Experimental Settings}


\subsubsection{Datasets}
In our experiments, we included both real and synthetic datasets.

(i)
For synthetic datasets, we use 90 synthetic objects and 27 test objects from the datasets provided by PU-GAN \cite{PUGAN}. For each training object, we use the same sampling setting as PU-Net \cite{PUNet}. We set the radius to 0.5, randomly select 100 patch pairs, and ensure that there is suitable overlaps between the pairs of patches. From these selected patches, we sample $n$ points as our training input and $rn$ points as our ground truth $\hat{Q}$ using Poisson sampling.
%
(ii) In order to verify the robustness of our method, we followed the same experimental setting as point cloud denoising network \cite{DMRDenoise,score-based,RePCD-Net}, and used the blensor \cite{blensor} to simulate real scan noise on the synthetic dataset.
(iii)
For the real datasets, we used ScanObjectNN \cite{uy-scanobjectnn-iccv19}, which contains 2902 point cloud objects in 15 categories; each object has 2048 points.

\subsubsection{Evaluation metrics}
We used three commonly-used metrics: Chamfer distance(CD), Point-to-Surface distance(P2F), and Hausdorff distance(HD).
The smaller value of these metrics are, the more effective the algorithm is.
\subsubsection{Comparison Methods}
To demonstrate the effectiveness of our network, we compared it with several state-of-the-art point cloud upsampling networks, including PU-Net \cite{PUNet}, PU-GAN \cite{PUGAN}, Dis-PU  \cite{dis-pu}, and PU-GCN  \cite{pugcn}. We used their open-source models for testing on the same test sets.
Note that, we did not compare it with PUGeo-Net \cite{PUGeo}, because it requires normals as an additional supervision during training, while the training dataset provided by PU-GAN \cite{PUGAN} lacks normals.
\begin{table}[b]
    \vspace{-0.4cm}
    \centering
    \footnotesize
    \caption{Quantitative comparisons between our method and state-of-the-arts. Note that CD, HD, and P2F are multiplied by 10$^3$. The best results are in bold. $\times4$ and $\times16$ indicate upsampling rate $r=4$, $r=16$, respectively.}
    \begin{tabular}{lcccc}
        \toprule
        Methods                   & \;\;\;CD\;\;\;  & \;\;\;HD\;\;\;  & P2F-mean        & P2F-std         \\
        \midrule
        \multicolumn{5}{l}{$\times4$ upsampling}                                                          \\
        PU-Net~\cite{PUNet}       & 0.5225          & 4.6083          & 4.8310          & 4.3590          \\
        PU-GAN~\cite{PUGAN}       & 0.2676          & 4.7379          & \textbf{1.9660} & 3.6100          \\
        PU-GCN~\cite{pugcn}       & 0.2724          & 3.0455          & 2.4780          & 3.3760          \\
        Dis-PU~\cite{dis-pu}      & 0.2560          & 4.7277          & 2.0230          & 3.1780          \\
        \textbf{PC$^2$-PU (ours)} & \textbf{0.2321} & \textbf{2.5942} & 2.1190          & \textbf{2.7250} \\
        \midrule
        \multicolumn{5}{l}{$\times16$ upsampling}                                                         \\
        PU-Net~\cite{PUNet}       & 0.3123          & 3.9111          & 5.9680          & 5.0470          \\
        PU-GAN~\cite{PUGAN}       & 0.2232          & 6.3243          & 2.5800          & 4.5690          \\
        PU-GCN~\cite{pugcn}       & 0.1657          & 3.8224          & 2.4370          & 3.5060          \\
        Dis-PU~\cite{dis-pu}      & 0.1484          & 6.0934          & \textbf{2.2620} & 3.7660          \\
        \textbf{PC$^2$-PU (ours)} & \textbf{0.0998} & \textbf{2.8692} & 2.3490          & \textbf{2.9440} \\
        \bottomrule
    \end{tabular}
    \label{benchmark_result}
\end{table}

\subsubsection{Implementation details}
In experiments, we set the point number of input patch $n$ = 256. We trained our network with a batch size of 32 for 400 epochs on the PyTorch platform. For fair comparison, we applied the experimental setup of PU-GAN \cite{PUGAN} for each patch, using random scaling, rotation, and point perturbation for data enhancement. The Adam optimizer is used with the learning rate of 0.001 at first training, which is linearly decreased by a decay rate of 0.1 per 20 epochs until $10^{-6}$.
Our training strategy followed \cite{dis-pu}, the parameter $\lambda$ in Eq \ref{rec} is linearly increased from 0.01 to 1.0 as the training progresses.
We will release the source code and the trained models upon the publication of this paper.

\subsection{Results on Synthetic Dataset}
We tested our network on the benchmark dataset provided by PU-GAN \cite{PUGAN}. Figure \ref{benchmark} shows the visual comparison on the benchmark dataset. Compared with the existing methods, our network effectively suppresses the generation of pseudo-points and maintains the spatial consistency efficiently.

The quantitative results with other advanced networks are shown in Table \ref{benchmark_result}. We can see that our method performed best in several important metrics, and the advantage of our method becomes more significant as the upsampling rate $r$ increases. The reason behind is that, existing methods are more likely to lose the spatial consistency of the target for larger upsampling rate, while our method achieves better performance by taking into account the relations from patch-to-patch and point-to-point.
\begin{figure}[h]
    \centering
    \includegraphics[width=1.0\columnwidth]{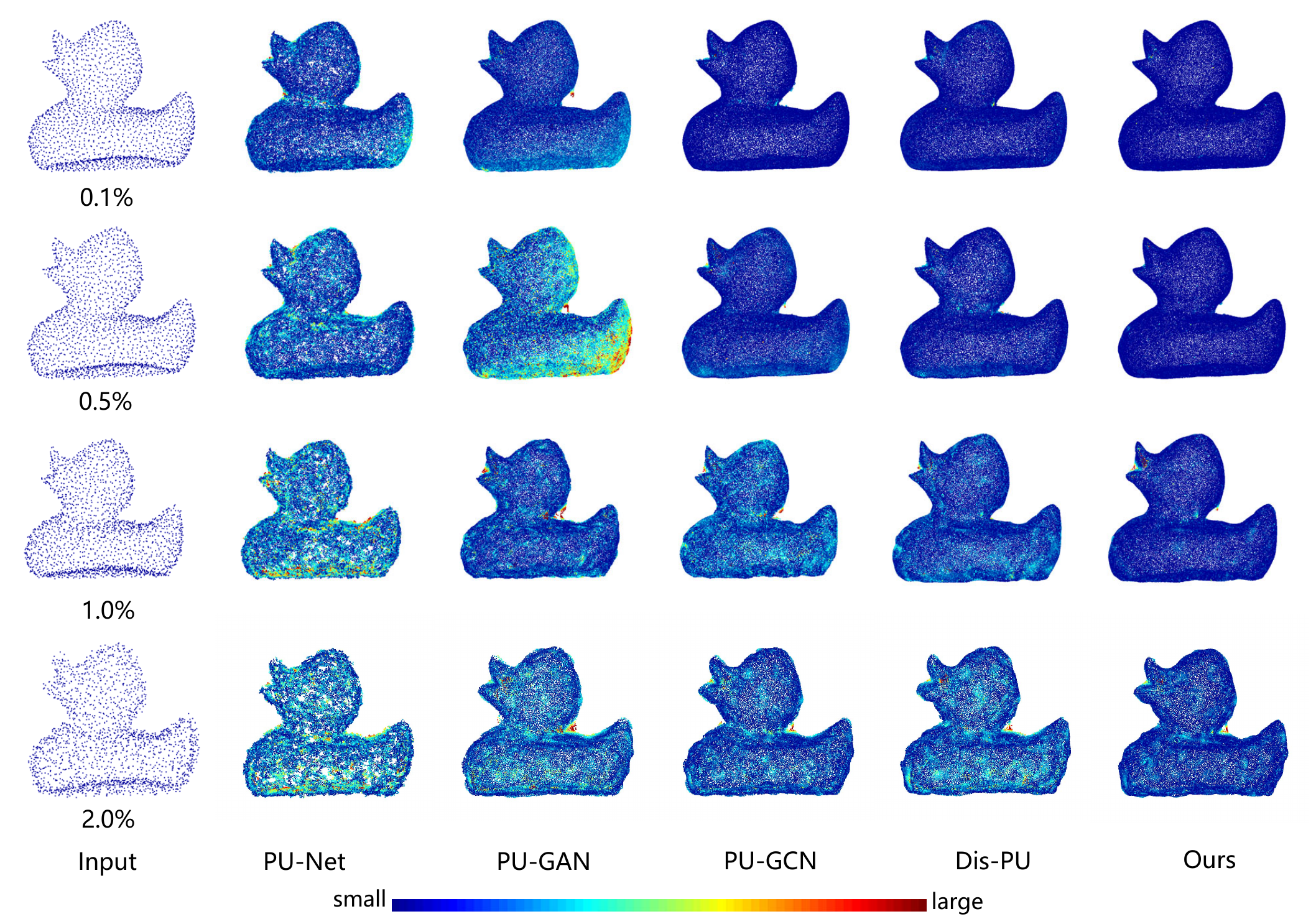} 
    \caption{Noise Robust Test with Gaussian noise visual comparison. We used different levels of Gaussian noise to compare our visual effect with other networks. The color indicates the CD value, and more blue means the better result.}
    \label{noise_1}
    \vspace{-0.4cm}
\end{figure}
\subsection{Results on real-scanned data}
We also conducted comparative experiments on the real dataset ScanobjectNN \cite{uy-scanobjectnn-iccv19}. As there is no ground truth, we only performed a visual comparison. We used the BallPivoting \cite{ballpivoting} algorithm to reconstruct the result after upsampling. Figure \ref{realscan} shows the visual comparison results. Compared with other networks, our network can preserve the underlying structure of the object so that the densified points are generated on the surface of the object as much as possible, such as the sofa back in Figure \ref{realscan}. Our results are smoother, without significant jitter, and with more fidelity in the overall structure. These experiments show that our network is still effective in maintaining the spatial consistency of the objects and improving the effectiveness of the network when applied to real scenarios.
\subsection{Robustness Test}
\begin{table}[h]
    \centering
    \footnotesize
    \caption{Robustness Test with Gaussian noise. We added Gaussian noise at 0.5\%, 1.0\%, and 2.0\% levels, and used CD as the evaluation metric multiplied by $10^{3}$. $\times4$ and $\times16$ indicate upsampling rate $r=4$, $r=16$, respectively.}
    \begin{tabular}{lcccc}
        \toprule
        Methods                   & 0.0\%           & 0.5\%           & 1.0\%           & 2.0\%           \\
        \midrule
        \multicolumn{5}{l}{$\times4$ upsampling}                                                          \\
        PU-Net~\cite{PUNet}       & 0.5225          & 0.5667          & 0.8724          & 1.3227          \\
        PU-GAN~\cite{PUGAN}       & 0.2676          & 0.3286          & 0.5114          & 0.9358          \\
        PU-GCN~\cite{pugcn}       & 0.2724          & 0.3011          & 0.4257          & 0.8339          \\
        Dis-PU~\cite{dis-pu}      & 0.2560          & 0.2792          & 0.4216          & 0.8464          \\
        \textbf{PC$^2$-PU (ours)} & \textbf{0.2321} & \textbf{0.2604} & \textbf{0.3586} & \textbf{0.7727} \\
        \midrule
        \multicolumn{5}{l}{$\times16$ upsampling}                                                         \\
        PU-Net~\cite{PUNet}       & 0.3123          & 0.3831          & 0.5048          & 0.8334          \\
        PU-GAN~\cite{PUGAN}       & 0.2232          & 0.2672          & 0.3935          & 0.7364          \\
        PU-GCN~\cite{pugcn}       & 0.1657          & 0.1937          & 0.3229          & 0.6777          \\
        Dis-PU~\cite{dis-pu}      & 0.1484          & 0.1748          & 0.3453          & 0.6904          \\
        \textbf{PC$^2$-PU (ours)} & \textbf{0.0998} & \textbf{0.1253} & \textbf{0.2163} & \textbf{0.5750} \\
        \bottomrule
    \end{tabular}
    \label{gaussian_noise}
\end{table}
\begin{table}[h]
    \centering
    \footnotesize
    \caption{Robustness Test with realistic LiDAR noise simulated by blensor. We set noise at 0.5\%, 1.0\%, and 2.0\% levels, and used CD as the evaluation metric in units of $10^{3}$. $\times4$ and $\times16$ indicate upsampling rate $r=4$, $r=16$, respectively.}
    \begin{tabular}{lcccc}
        \toprule
        Methods                   & 0.5\%           & 1.0\%           & 2.0\%           \\
        \midrule
        \multicolumn{4}{l}{$\times4$ upsampling}                                        \\
        PU-Net~\cite{PUNet}       & 0.7122          & 0.9388          & 2.0315          \\
        PU-GAN~\cite{PUGAN}       & 0.3587          & 0.6812          & 1.6510          \\
        PU-GCN~\cite{pugcn}       & 0.3279          & 0.5505          & 1.6049          \\
        Dis-PU~\cite{dis-pu}      & 0.3114          & 0.5804          & 1.5874          \\
        \textbf{PC$^2$-PU (ours)} & \textbf{0.2773} & \textbf{0.4233} & \textbf{1.3964} \\
        \midrule
        \multicolumn{4}{l}{$\times16$ upsampling}                                       \\
        PU-Net~\cite{PUNet}       & 0.3631          & 0.5197          & 1.2855          \\
        PU-GAN~\cite{PUGAN}       & 0.2292          & 0.5078          & 1.4140          \\
        PU-GCN~\cite{pugcn}       & 0.1655          & 0.3966          & 1.4184          \\
        Dis-PU~\cite{dis-pu}      & 0.1555          & 0.4212          & 1.3702          \\
        \textbf{PC$^2$-PU (ours)} & \textbf{0.1008} & \textbf{0.2474} & \textbf{1.1182} \\
        \bottomrule
    \end{tabular}
    \label{blensor_noise}
\end{table}
Inspired by point cloud denoising methods \cite{DMRDenoise,score-based}, we design the following experiments to demonstrate the robustness of our network in handling noise.

We first used the benchmark dataset provided by PU-GAN and added different levels of Gaussian noise to the normalized objects at each object. We added 0.5\%, 1.0\%, and 2.0\% noise, respectively, with the corresponding quantitative evaluation shown in Table \ref{gaussian_noise}. We can see that our network shows its advantages in handling noise, and the gap between ours and other methods increases as the noise level increases. Figure \ref{noise_1} shows our visual comparison, and the color of the dots in the figure represents the CD to the ground truth. Compared to other networks, when the noise increases significantly, we retain the local consistency of the target better.

Besides, we randomly selected ten objects as models from the benchmark dataset provided by PU-GAN and simulated the scanning scenario using the blensor \cite{blensor} plugin. Compared to directly adding Gaussian noise, this simulation of realistic scanning method can produce more realistic point clouds and noise. The quantitative results are shown in Table \ref{blensor_noise}. Obviously, our network still had a significant improvement over the existing networks when faced with realistic noise.

\subsection{Ablation Study}
\subsubsection{Different Modules in the Framework.}
To demonstrate the effectiveness of each component proposed in our network, we performed the corresponding ablation experiments by removing each component in the following cases.
\begin{itemize}
    \item Remove the Patch Correlation Module.
    \item Remove the Point Correlation Module.
    \item Remove the SPNE and LSE by using the 3D coordinates as the position code of the point.
\end{itemize}
\begin{table}[h]
    \centering
    \footnotesize
    \caption{Ablation study. The evaluation results of different components of PC$^2$-PU. We used CD as a quantitative evaluation metric multiplied by $10^{3}$. $\times4$ and $\times16$ indicate upsampling rate $r=4$, $r=16$, respectively.}
    \begin{tabular}{lcccc}
        \toprule
        Methods                      & 0.0\%           & 0.5\%           & 1.0\%           \\
        \midrule
        \multicolumn{4}{l}{$\times4$ upsampling}                                           \\
        w/o patch correlation module & 0.2495          & 0.2801          & 0.3846          \\
        w/o point correlation module & 0.2425          & 0.2632          & 0.3713          \\
        w/o position encoder         & 0.2612          & 0.2760          & 0.3987          \\
        \textbf{Full Model}          & \textbf{0.2321} & \textbf{0.2604} & \textbf{0.3586} \\
        \midrule
        \multicolumn{4}{l}{$\times16$ upsampling}                                          \\
        w/o patch correlation module & 0.1275          & 0.1525          & 0.2713          \\
        w/o point correlation module & 0.1153          & 0.1348          & 0.2359          \\
        w/o position encoder         & 0.1371          & 0.1760          & 0.3108          \\
        \textbf{Full Model}          & \textbf{0.0998} & \textbf{0.1253} & \textbf{0.2163} \\
        \bottomrule
    \end{tabular}
    \label{ablation_table}
\end{table}
We added different levels of Gaussian noise to the benchmark dataset \cite{PUGAN} for evaluation, and the experimental results are shown in the Table \ref{ablation_table}.

Obviously, our full pipeline gets the minimum CD value for the experiments, removing any component degrades the overall performance of the network, which means that every component in our framework contributes to the upsampling task. In addition, it is worth noting that, \emph{after removing the proposed SPNE and LSE, the network achieved the worst results in the ablation experiment}. This is because the main idea of our proposed method PC$^2$-PU is to exploit the point-to-point and patch-to-patch relations to achieve more efficient and robust upsampling. Removing the SPNE and LSE will lose these relationships, thus limiting the effectiveness of the upsampling.

\subsubsection{Adjacent patches vs. Enlarging patch size.}

\begin{table}[h]
    \centering
    \footnotesize
    \caption{Comparison with increasing patch size. Note that CD, HD, and P2F are multiplied by $10^{3}$. The best results are in bold.
        $\times4$ indicates upsampling rate $r=4$.
    }
    \begin{tabular}{lcccc}
        \toprule
        Methods                   & \;\;\;CD\;\;\;  & \;\;\;HD\;\;\;  & P2F-mean        & P2F-std         \\
        \midrule
        \multicolumn{5}{l}{$\times4$ upsampling}                                                          \\
        patch size(x2)            & 0.2528          & 2.7056          & 2.2880          & 3.0450          \\
        patch size(x4)            & 0.2746          & 2.6058          & 2.8410          & 3.2220          \\
        AG-GCN~\cite{AGGAN}       & 0.3993          & 3.6426          & 4.9220          & 5.3332          \\
        Dis-PU~\cite{dis-pu}      & 0.2560          & 4.7277          & 2.0230          & 3.1780          \\
        \textbf{PC$^2$-PU (ours)} & \textbf{0.2321} & \textbf{2.5942} & \textbf{2.1190} & \textbf{2.7250} \\
        \bottomrule
    \end{tabular}
    \label{big_patch}
\end{table}
\begin{table}[t]
    \centering
    \footnotesize
    \caption{Generalization Ability Test. Note that CD, HD, and P2F are multiplied by $10^{3}$. The best results are in bold. $\times4$ indicates upsampling rate $r=4$.}
    \begin{tabular}{lcccc}
        \toprule
        Methods              & \;\;\;CD\;\;\;  & \;\;\;HD\;\;\;  & P2F-mean        & P2F-std         \\
        \midrule
        \multicolumn{5}{l}{$\times4$ upsampling}                                                     \\
        patch size(x2)       & 0.5594          & 6.1717          & 1.7450          & 2.6970          \\
        patch size(x4)       & 0.5701          & 6.0004          & 1.8750          & 2.5410          \\
        AG-GCN~\cite{AGGAN}  & 0.7185          & 7.4686          & 2.3390          & 2.6670          \\
        Dis-PU~\cite{dis-pu} & 0.5339          & 6.0474          & 1.7520          & 2.9250          \\
        PC$^2$-PU (ours)     & \textbf{0.5099} & \textbf{5.9618} & \textbf{1.4940} & \textbf{2.2190} \\
        \bottomrule
    \end{tabular}
    \label{g_ability}
\end{table}

We have mentioned in Section \ref{PRU} that,
instead of increasing the patch size, we designed PaCM to capture the relationship between adjacent patches. To validate our design, we retrained the network, enlarging the patch size with $\times2$(512 points) and $\times4$(1024 points), and replaced the SPNE in \emph{Patch Correlation Module} with LSE. Besides, we compared with AG-GCN \cite{AGGAN}, which uses the complete target as input. The corresponding experimental results are shown in the Table \ref{big_patch}.

To further validate the generalization of our method, we directly leverage our trained model on PU1K models provided by PU-GCN \cite{pugcn}.
The experimental results are shown in the Table \ref{g_ability}. We can see that increasing the patch size does not improve the upsampling performance, while the generalization ability of the model decreases as the patch size increases.
\subsubsection{Model Complexity}
Unlike some networks that use transformer, our network is a lightweight point cloud upsampling network. To prove that, we compared the model complexity of the existing methods, and the results are shown in the Table \ref{model_complex}.
\begin{table}[h]
    \centering
    \footnotesize
    \caption{Model complexity comparison. Note the CD is multiplied by $10^{3}$. The best results are in bold, and second best results are underlined.}
    \begin{tabular}{lcccc}
        \toprule
        Methods                   & Trainable params     & FLOPs                & Inference time       & CD                 \\
        \midrule
        PU-Net\cite{PUNet}        & 3.0000MB             & 1.0000GB             & \underline{4.5530ms} & 0.5225             \\
        PU-GAN\cite{PUGAN}        & 2.0660MB             & \underline{0.9074GB} & 7.0788ms             & 0.2676             \\
        PU-GCN\cite{pugcn}        & \textbf{0.2898MB}    & \textbf{0.3841GB}    & 5.5062ms             & 0.2724             \\
        Dis-PU\cite{dis-pu}       & 3.9939MB             & 3.0519GB             & 12.7729ms            & \underline{0.2560} \\
        \textbf{PC$^2$-PU (ours)} & \underline{1.7102MB} & 1.3470GB             & \textbf{4.1867ms}    & \textbf{0.2321}    \\
        \bottomrule
    \end{tabular}
    \label{model_complex}
\end{table}

The number of trainable parameters in our proposed PC$^2$-PU is second only to PU-GCN \cite{pugcn}, and only half of the current state-of-the-art method Dis-PU \cite{dis-pu}.
Moreover, our network achieves a good advantage in inference time, because we only perform transformer operation in the local neighborhood of points \cite{zhao-point}, which can utilize GPU to effectively accelerate the transformer structure \cite{Nico-point, fan2017point} without introducing extra parameters.
Overall, PC$^2$-PU is a powerful and affordable model for point cloud upsampling tasks.

\section{Conclusion}

In this paper, we present a novel method for effective and robust point cloud upsampling.
Different from existing methods that upsampling each patch separately, we propose to leverage adjacent patch to amend the lost geometry information in a single patch.
Our key idea is to leverage patch-to-patch and point-to-point relations to enhance the spatial consistency of the upsampled surface.
To this end, we formulate a novel network PC$^2$-PU that consist of a \emph{Patch Correlation Module} to combine the relationship between adjacent patches, and a \emph{Point Correlation Module} to maintain the local spatial consistency by revealing the relationship between points.
Extensive comparative experiments show that our proposed network outperform current SOTA methods on both clean and noisy inputs.

There are also some limitations to this work. For example, since our method uses patch pairs as input, the patch selection strategy can be directly related to the stability of the algorithm. We will explore more stable and reasonable ways to exploit these relationships in future research.
\begin{acks}
This work was supported by the National Natural Science Foundation of China under Grant 42172431.
\end{acks}

\bibliographystyle{ACM-Reference-Format}
\bibliography{ref}

\appendix
\section{Supplementary experiment}
\subsection{Patch Selection} \label{a1}
Unlike existing patch-based approaches, the input of our network is a pair of adjacent patches.

To select patches during testing, we adopt a patch selection strategy, which is detailed in the Algorithm \ref{alg:algorithm}. Specifically, we first select the three closest patches at each patch, and then use density clustering \cite{ester1996density} to find the patch with the highest number of categories in the overlapping region of the patch as an additional input. In this way, we are able to find pairs of patches with richer geometric structure as inputs.

\begin{algorithm}[h]
    \caption{Adjacent patch pair selection method} \label{alg:algorithm}
    \begin{algorithmic}[1] 
        \STATE \textbf{Input}: low-resolution patch sets $PS = \{P_1, P_2 .... P_M\}$
        \STATE \textbf{Output}: Pairs of patch sets $PPS = \{(P_1, P_1^{opt}),...(P_M, P_M^{opt})\}$
        \FOR{$P_m$ in $PS$}
        \STATE Select the $K$ closest patches to $P_m$ based on distance $PN = \{P_m^1, P_m^2 ... P_m^K\}, PN \in PS$
        \STATE Get overlapping area between $P_m$ and its neighboring patches. $OA = \{P_m \cap P_m^1, ... P_m \cap P_m^K\}$
        \FOR{$OA_k = P_m \cap P_m^K$ in $OA$}
        \STATE
        Apply density clustering algorithm \cite{ester1996density} to cluster $QA_k$ and get the number of clusters $C_k$.
        \ENDFOR
        \STATE $opt = {\underset {k} {argmax}}(C_k), k \in \{1, 2, ... K\}$.
        \STATE Push $(P_m, P_m^{opt})$ to $PPS$
        \ENDFOR
    \end{algorithmic}
\end{algorithm}

\subsection{Comparison with Denoise method}
We used a combination of denoised and upsampled networks to compare with our network to demonstrate the superiority of our network. We used the current state-of-the-art method Score-based Denoising \cite{score-based} as the point cloud denoising network. First used the Score-based denoising to denoise the dataset with Gaussian noise, and then used PU-Net, PU-GAN, etc network for upsampling before comparing the corresponding results with our network. It is worth noting that, for fair comparison with denoising networks, we retrained PC$^2$-PU using the training strategy proposed by score-based. Unlike existing methods that use 1.0\% Gaussian noise as data augmentation during training, we randomly select 0-2.0\% noise to perturb the input. Such a training method reduced the upsampling effect of the network due to the instability of the input, but improved the robustness of the network instead. The corresponding quantitative evaluation results are shown in Table \ref{denosie}.

\begin{table}[ht]
    \centering
    \scriptsize
    \caption{Comparison with point cloud denoising network. We added different levels of Gaussian noise, and compared our method with other methods. We used CD as a quantitative evaluation metric multiplied by $10^{3}$.}
    \begin{tabular}{lccc}
        \toprule
        Methods                                      & 0.5\%           & 1.0\%           & 2.0\%           \\
        \midrule
        \multicolumn{4}{l}{$\times4$ upsampling}                                                           \\
        Score\cite{score-based}+PU-Net~\cite{PUNet}  & 0.7234          & 0.7580          & 0.7968          \\
        Score\cite{score-based}+PU-GAN~\cite{PUGAN}  & 0.5023          & 0.5216          & 0.6466          \\
        Score\cite{score-based}+PU-GCN~\cite{pugcn}  & 0.4849          & 0.5176          & 0.6125          \\
        Score\cite{score-based}+Dis-PU~\cite{dis-pu} & 0.4590          & 0.5061          & 0.6000          \\
        \textbf{PC$^2$-PU (ours)}                    & \textbf{0.3237} & \textbf{0.3535} & \textbf{0.5885} \\
        \midrule
        \multicolumn{4}{l}{$\times16$ upsampling}                                                          \\
        Score\cite{score-based}+PU-Net~\cite{PUNet}  & 0.5729          & 0.5658          & 0.6028          \\
        Score\cite{score-based}+PU-GAN~\cite{PUGAN}  & 0.3918          & 0.4070          & 0.5344          \\
        Score\cite{score-based}+PU-GCN~\cite{pugcn}  & 0.3731          & 0.3916          & 0.4748          \\
        Score\cite{score-based}+Dis-PU~\cite{dis-pu} & 0.3337          & 0.3823          & 0.4746          \\
        \textbf{PC$^2$-PU (ours)}                    & \textbf{0.2370} & \textbf{0.2543} & \textbf{0.4727} \\
        \bottomrule
    \end{tabular}
    \label{denosie}
\end{table}
Compared with this denoising method, our network had a clear advantage when the noise level is small. This is due to the fact that, most existing point cloud denoising networks are trained on datasets with a large number of points like ShapeNet  \cite{chang2015shapenet}, so they are not effective in denoising when applied to sparse upsampled datasets. Also, the denoising network is more likely to lose sparse details when the noise level is low.

\section{Visual Comparison Experiment}
As shown in the Fig \ref{denosie}, we added 1.0\% Gaussian noise and compared our network with other networks. We used colors to indicate the magnitude of the error, and our network has a clear advantage over the other networks.

\section{Comparison of coarse output and refined output}
To illustrate the specific role of the Point Correlation Module, we performed a visual comparison of the rough point cloud $Q^\prime$ with the refined point cloud $Q$ after the Point Correlation Module. We first added different levels of Gaussian noise to the dataset, and the specific experimental results are shown in Fig \ref{ablation_study}. We can see from the Fig \ref{ablation_study} that the error of the point set is significantly reduced after the Point Correlation Module, which effectively improves the the performance of upsampling.
\begin{figure*}[h]
    \centering
    \includegraphics[width=2.0\columnwidth]{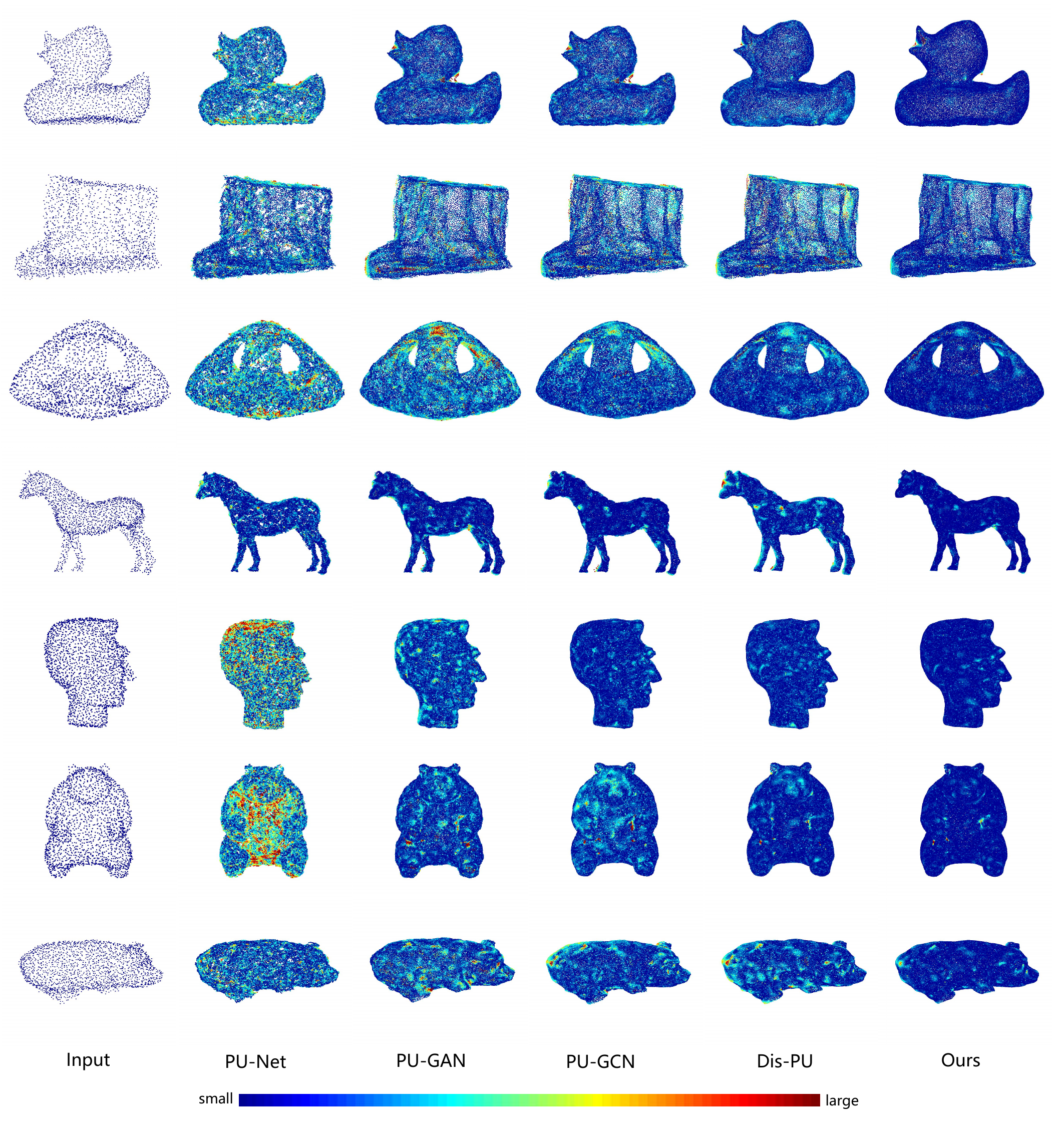} 
    \caption{Visual comparison on noise robust test with gaussian noise. We used a Gaussian noise of 1.0\% as an example to compare our visual effect with other networks. Where the colors reveal the CD distance, and the smaller the CD distance, the better the result.}
    \label{noise_2}
\end{figure*}
\begin{figure*}[h]
    \centering
    \includegraphics[width=2.0\columnwidth]{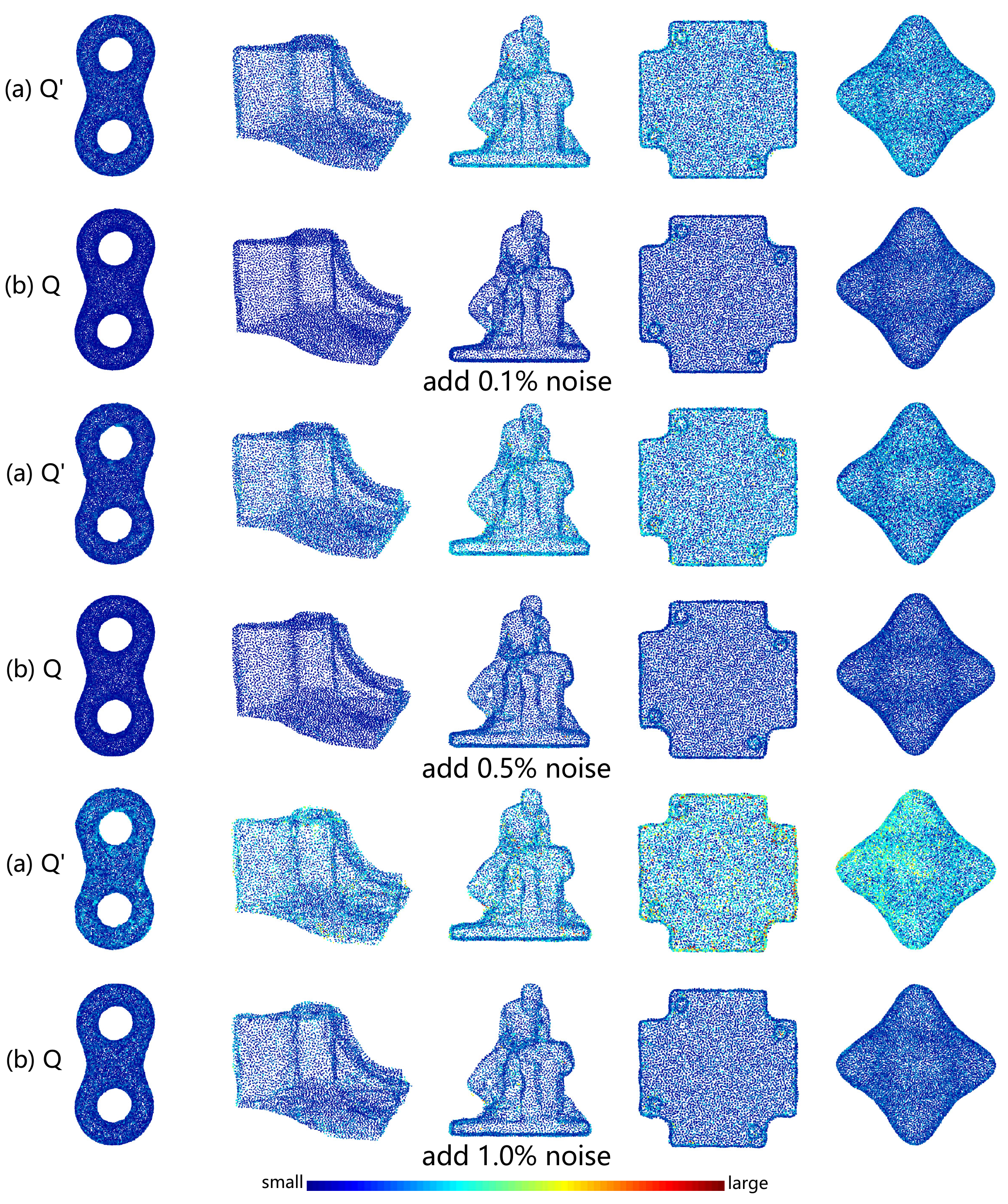} 
    \caption{Visualization results for the ablation study. (a) denotes our rough output $Q^\prime$ and (b) denotes the corrected result $Q$ after the Point Correlation Module. Where the colors reveal the CD distance, and the smaller the CD distance, the better the result.}
    \label{ablation_study}
\end{figure*}

\end{document}